\documentclass[10pt,twocolumn,letterpaper]{article}

\usepackage{cvpr}              % To produce the CAMERA-READY version
\definecolor{cvprblue}{rgb}{0.21,0.49,0.74}
\usepackage[pagebackref,breaklinks,colorlinks,allcolors=cvprblue]{hyperref}
\usepackage{amsmath,amssymb,mathtools}
\usepackage{algorithm}
\usepackage{algpseudocode}
\usepackage{multirow}
\usepackage{booktabs}
\usepackage{arydshln}
\usepackage{url}
\usepackage{stfloats}

\def\paperID{} % *** Enter the Paper ID here
\def\confName{CVPR}
\def\confYear{2026}

\title{SenCache: Accelerating Diffusion Model Inference via \\Sensitivity-Aware Caching}

\author{
Yasaman Haghighi \qquad Alexandre Alahi\\
École Polytechnique Fédérale de Lausanne (EPFL)\\
{\tt\small yasaman.haghighi@epfl.ch \quad alexandre.alahi@epfl.ch}
}

\begin{document}
\twocolumn[{%
\renewcommand\twocolumn[1][]{#1}%
%\vspace{-0.5cm}
\maketitle
% % Remove page # from the first page of camera-ready.
% % \ificcvfinal\thispagestyle{empty}\fi

% % Under the same computational budget, SenCache improves the visual quality of the generated samples compared to previous caching strategies

\begin{center}
  \newcommand{\teaserwidth}{\textwidth}
  \vspace{-0.3cm}
  \centerline{\includegraphics[width=\linewidth]{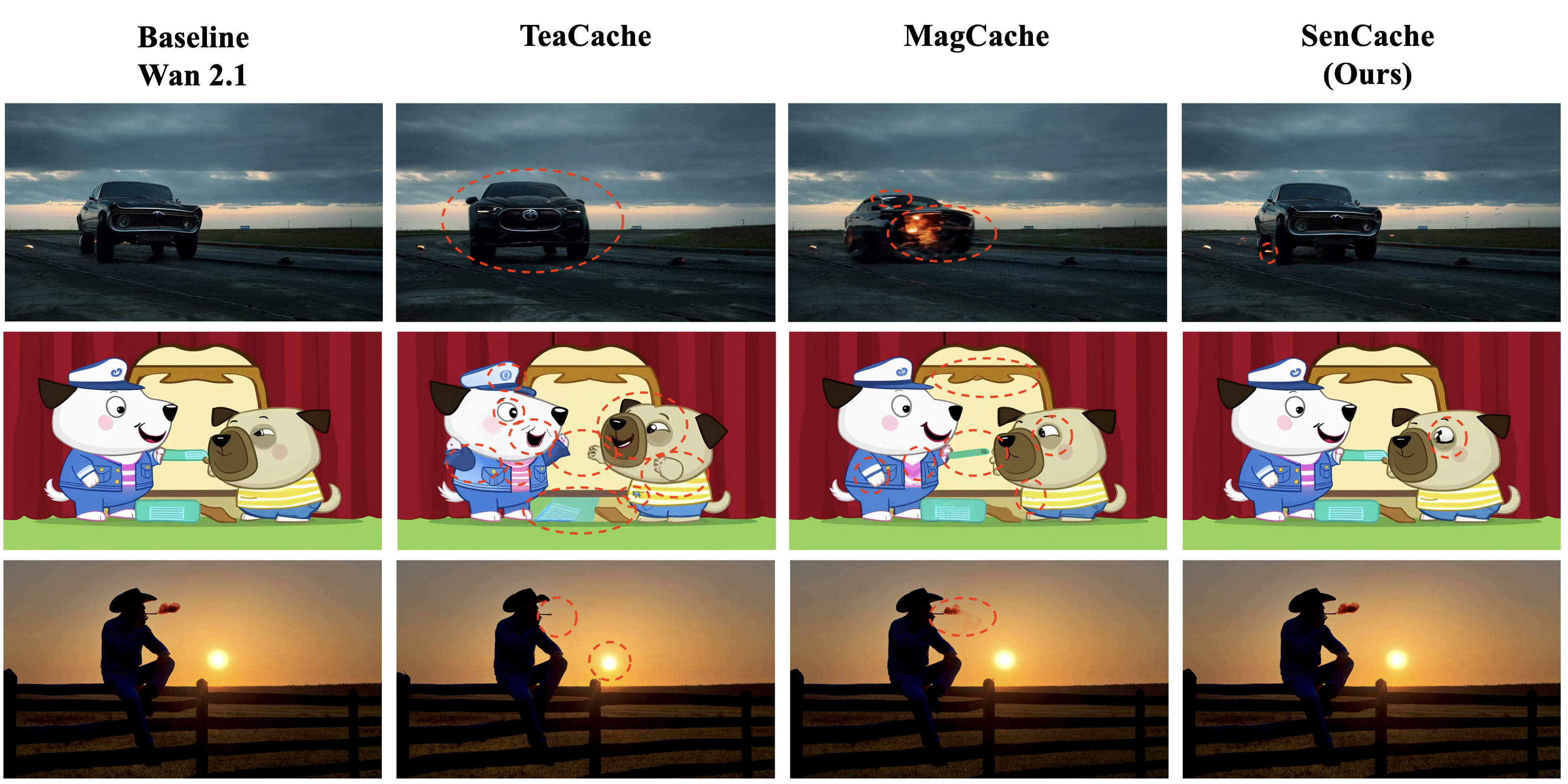}}
    \captionof{figure}{\textbf{SenCache} is a caching algorithm for accelerating the inference of diffusion models. Unlike prior methods that rely on heuristics, SenCache uses a theoretically motivated measure of network sensitivity to its input perturbations as the criterion for caching. All examples are generated with Wan 2.1~\cite{wan2025open}. Under the same compute budget, SenCache better preserves the visual quality of the generated samples.}
    \label{fig:teaser}
\end{center}%
}]
\begin{abstract}
Diffusion models achieve state-of-the-art video generation quality, but their inference remains expensive due to the large number of sequential denoising steps. This has motivated a growing line of research on accelerating diffusion inference. Among training-free acceleration methods, caching reduces computation by reusing previously computed model outputs across timesteps. Existing caching methods rely on heuristic criteria to choose cache/reuse timesteps and require extensive tuning. We address this limitation with a principled sensitivity-aware caching framework. Specifically, we formalize the caching error through an analysis of the model output sensitivity to perturbations in the denoising inputs, i.e., the noisy latent and the timestep, and show that this sensitivity is a key predictor of caching error. Based on this analysis, we propose Sensitivity-Aware Caching (\textbf{SenCache}), a dynamic caching policy that adaptively selects caching timesteps on a per-sample basis. Our framework provides a theoretical basis for adaptive caching, explains why prior empirical heuristics can be partially effective, and extends them to a dynamic, sample-specific approach. Experiments on Wan 2.1, CogVideoX, and LTX-Video show that SenCache achieves better visual quality than existing caching methods under similar computational budgets. The code is available at \small \href{https://github.com/vita-epfl/SenCache.git}{\texttt{https://github.com/vita-epfl/SenCache.git}}
\end{abstract}    
\section{Introduction}
\label{sec:intro}
Diffusion models~\citep{ho2020ddpm,song2021sde} and flow matching models ~\cite{albergo2022building,lipman2022flow} have reshaped generative modeling by setting the state of the art in image and video synthesis. Despite their success, diffusion inference remains computationally expensive: sample generation requires multiple denoising iterations, and each iteration incurs a full forward pass of a large network. This cost is especially prohibitive for modern video diffusion transformers, which contain billions of parameters and can require minutes of computation even for a few seconds of video~\cite{wan2025open, yang2025cogvideox}. Reducing inference latency—without retraining the model or degrading output quality—has therefore become a key challenge for practical deployment.

Among acceleration strategies, caching-based methods~\cite{Kahatapitiya2025AdaCache, liu2025teacache, ma2025magcache} are particularly appealing because they reduce inference cost by reusing previously computed denoiser outputs, without retraining (as in distillation-based methods ~\citep{Salimans2022PD,Song2023Consistency,Luo2023LCM}) or modifying the model architecture. The underlying premise is that denoiser outputs at consecutive timesteps can be sufficiently similar, allowing cached outputs to replace expensive forward evaluations. Existing methods, however, determine these reuse timesteps using empirical heuristics. For instance, TeaCache~\citep{liu2025teacache} builds cache-reuse rules through output residual modeling with time embedding difference or modulated input difference, while MagCache~\citep{ma2025magcache} selects reuse timesteps based on the magnitude of the residual (the difference between the model’s prediction and its input). While effective in favorable regimes, these heuristics have two fundamental limitations: (1) they lack theoretical justification and require extensive hyperparameter tuning, and (2) they produce static caching schedules that cannot adapt to the varying difficulty of each sample. As a consequence, caching may over-cache challenging samples or under-cache easy ones, because reuse decisions are not adapted to sample-specific dynamics.

In this work, we propose a sensitivity-based criterion for cache/reuse decisions in diffusion inference. Our key idea is to use the denoiser’s local sensitivity—i.e., the variation of its output with respect to perturbations in the noisy latent and timestep—as a proxy for output change between neighboring denoising steps. We show that this variation is well characterized by the denoiser’s derivatives with respect to the noisy latent and the timestep. These sensitivities quantify the effect of latent drift and timestep spacing on the denoiser output, allowing us to predict when the induced output change is sufficiently small for cache reuse. 

Through analysis and empirical study, we show that local sensitivity is a strong predictor of caching error, and that both latent and timestep sensitivities contribute significantly. This reveals a core limitation of prior heuristic policies, which do not explicitly model both sources of variation.

Motivated by this insight, we introduce \textbf{Sensitivity-Aware Caching (SenCache)}, a principled, dynamic caching framework that adapts cache/reuse decisions to each sample. At every denoising step, SenCache predicts the denoiser output change using a first-order sensitivity approximation and reuses the cached output only when the predicted deviation is below a target tolerance. Our experiments on three state of the art video diffusion models, Wan 2.1 \cite{wan2025open}, CogVideoX \cite{yang2025cogvideox}, and LTX-Video \cite{HaCohen2024LTXVideo}, show that SenCache outperforms existing caching strategies in visual quality under similar computational budgets. 

SenCache framework offers several advantages:
\begin{enumerate}
\item It provides a theoretically motivated decision rule for caching with an explicit tolerance for controlling the speed–quality trade-off.
\item It provides a sensitivity-based interpretation of why prior heuristics succeed in some regions and fail in others. 
\item It adapts cache/reuse decisions per sample, unlike prior methods that use fixed timesteps for all samples.
\item It requires no additional training and no model modification, and is agnostic to architecture and sampler.
\item While our experiments focus on visual domain, the underlying principle of using network sensitivity as a proxy for cache/reuse decisions is general and can be extended to other domains, such as audio and human motion.
\end{enumerate}

\section{Related Work}
\label{sec:relatedwork}
Diffusion models~\citep{ho2020ddpm,song2021sde} and flow matching models ~\cite{albergo2022building,lipman2022flow} have become a foundational tool for high-quality image and video synthesis~\citep{wan2025open,yang2025cogvideox,HaCohen2024LTXVideo,ma2024sit}. Early video diffusion systems extended 2D U\hbox{-}Net--based image models to the temporal domain~\citep{ho2022vdm,ho2022imagenvideo,singer2022makeavideo,zhou2022magicvideo}, but the limited receptive field of U\hbox{-}Nets makes it difficult to model long-range spatiotemporal dependencies. This motivated the introduction of Diffusion Transformers (DiTs)~\citep{peebles2023dit}, which now form the backbone of many state of the art text-to-video generators. Large-scale systems such as CogVideoX~\citep{yang2025cogvideox} and Wan~2.1~\citep{wan2025open} adopt DiTs with expert transformer modules and deliver strong visual quality, but producing a 5-second clip can still take several minutes on a single A800 GPU. LTX-Video~\citep{HaCohen2024LTXVideo} further improves efficiency by tightly coupling a Video-VAE with a DiT-based denoiser. Despite these advances, current DiT-based video generators remain computationally expensive, underscoring the need for faster inference methods.

\paragraph{Reducing Per-Step Cost: Quantization, Pruning, and NAS.} One strategy for accelerating diffusion inference is to reduce the computation of each denoising step. Quantization reduces precision while attempting to preserve fidelity via post-training calibration or light finetuning but typically needs task-specific calibration data and care to avoid timestep-wise error accumulation~\citep{Shang2023PTQ4DM,Li2023QDiffusion,Li2023QDM,viditq2024}. Pruning removes channels/blocks to shrink FLOPs, yet commonly entails additional optimization or data-dependent criteria to retain quality~\citep{Fang2023DiffPruning,Castells2024LDPruner}. A complementary line uses (training-free or lightly supervised) neural architecture search to co-design timesteps and lighter backbones, but the search still incurs non-trivial compute and may require workload-specific tuning~\citep{Li2023AutoDiffusion,Flexiffusion2025}.

\paragraph{Reducing the Number of Sampling Steps.} 
Distillation methods explicitly learn few-step generators: progressive distillation halves steps iteratively~\citep{Salimans2022PD}, while Consistency Models and their latent variants (LCM) directly learn mappings that support 1–4 step generation~\citep{Song2023Consistency,Luo2023LCM}. These approaches substantially reduce step counts but typically demand additional training and can be sensitive to domain and guidance settings.

\paragraph{Caching-Based Acceleration.} Caching approaches accelerate diffusion inference by reusing computations across timesteps. For U-Net models, DeepCache reuses high-level features across adjacent timesteps to cut redundant work with minimal quality loss~\citep{Ma2024DeepCache}. For DiT-style transformers, $\Delta$-DiT caches residuals between attention layers to tailor caching to transformer blocks~\citep{chen2024delta}, while FORA reuses intermediate attention/MLP outputs across steps without retraining~\citep{Selvaraju2024FORA}. PAB targets video DiTs by broadcasting attention maps in a pyramid schedule, exploiting the U-shaped redundancy of attention differences to reach real-time generation~\citep{Zhao2024PAB}. Beyond fixed schedules, AdaCache adapts caching decisions to content/timestep dynamics for video DiTs~\citep{Kahatapitiya2025AdaCache}. FasterCache further shows strong redundancy between conditional and unconditional branches in classifier-free guidance and reuses them efficiently within a timestep~\citep{Lv2024FasterCache}. Some methods require learning an explicit caching router (e.g., Learning-to-Cache for DiTs), adding optimization overhead~\citep{Ma2024L2C}; others are training-free but may still need calibration to avoid cumulative errors on long videos.

\paragraph{Full-Forward Caching.} Another family of methods performs \emph{full-forward caching}, storing the denoiser network outputs at selected timesteps rather than intermediate features. Recent examples of such methods are TeaCache and MagCache~\citep{liu2025teacache,ma2025magcache} and concurrent work LeMiCa~\citep{gao2025lemica}.  TeaCache builds prompt-specific skipping rules based on residual modeling, which risks overfitting and requires nontrivial calibration. MagCache uses residual magnitude heuristics and assumes a consistent ``magnitude law" across models and prompts. LeMiCa takes a different perspective and formulates cache scheduling as a global path optimization problem (lexicographic minimax) to control worst-case accumulated error across steps. While effective under well-tuned settings, these approaches rely on heuristic triggers without theoretical guarantees and often require extensive hyperparameter tuning or optimization to balance speed and quality.

\paragraph{Differences with Previous Methods.}
SenCache is also a full-forward caching approach, but it replaces ad-hoc heuristics with cache decisions grounded in the denoiser's local sensitivity to $\mathbf{x}_t$ and $t$. Specifically, SenCache caches only when a first-order sensitivity-based estimate predicts that the output change is small. This criterion is modality-agnostic, architecture-agnostic, and sampler-agnostic, since it depends on local model sensitivity and the actual input change between steps, rather than on hand-crafted triggers. As a result, the framework extends naturally across settings where heuristic rules may fail when their assumptions do not hold.

% \section{Methodology}
% \label{sec:method}
\begin{figure*}[t]
    \centering
    \begin{subfigure}[b]{\textwidth}
        \centering
        \includegraphics[width=.75\textwidth]{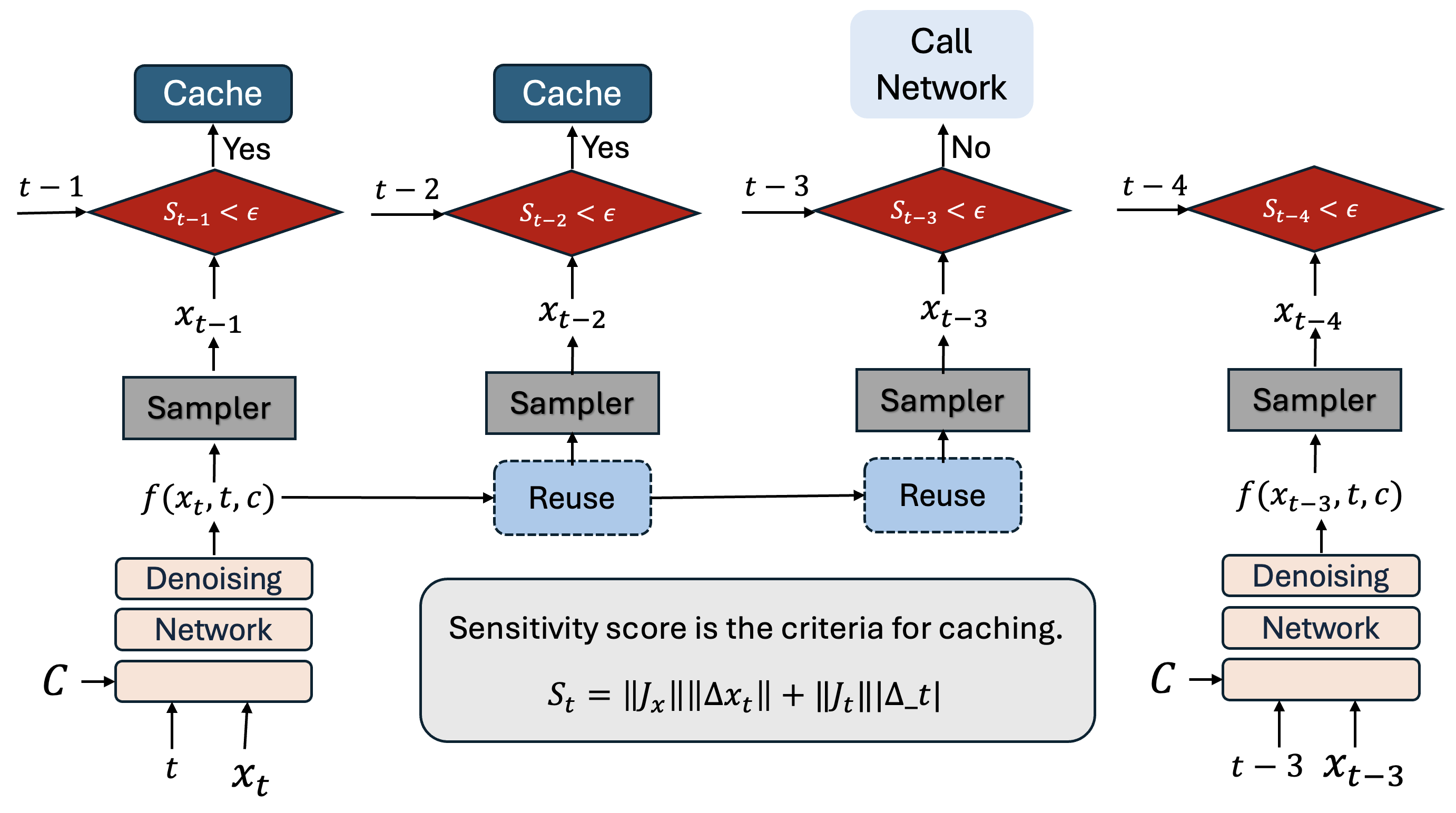}
    \end{subfigure}
    \caption{\textbf{SenCache uses sensitivity as a caching criterion.} At each denoising step, if the changes in the noisy latent $x_t$ and the sampling step $t$ are sufficiently small such that the sensitivity score (see \Cref{eq:sensitivity-score}) falls below $\varepsilon$, we reuse the cached denoiser output; otherwise, we refresh the cache at the current state. By skipping expensive denoiser evaluations when the output is expected to change minimally, SenCache accelerates diffusion-model inference.
    }
    \label{fig:pipelien}
\end{figure*}
\section{Background}

\paragraph{Flow Matching and the Probability Flow ODE.}
We adopt the flow-matching view of diffusion models \cite{albergo2023stochastic,ma2024sit}, where a data sample $\mathbf{x}_0 \sim p_{\mathrm{data}}$ is continuously transformed into a noisy variable $\mathbf{x}_t$ over $t \in [0,T]$ via
\begin{equation}
\mathbf{x}_t = \alpha_t \mathbf{x}_0 + \sigma_t \boldsymbol{\epsilon}, \qquad \boldsymbol{\epsilon} \sim \mathcal{N}(\mathbf{0},\mathbf{I}),
\end{equation}
where $\alpha_t$ and $\sigma_t$ are scalar schedules with $\alpha_0=1,\ \sigma_0=0$ and $\alpha_T=0,\ \sigma_T=1$.

The marginal distribution of $\mathbf{x}_t$ evolves under a velocity field $\mathbf{v}(\mathbf{x},t)$, and sampling follows the associated ODE
\begin{equation}
\dot{\mathbf{x}}_t = \mathbf{v}(\mathbf{x}_t,t).
\label{ode-eq}
\end{equation}
For the interpolation above, the conditional target velocity is
\begin{equation}
\mathbf{v}(\mathbf{z},t)
= \mathbb{E}\!\left[\dot{\alpha}_t\mathbf{x}_0+\dot{\sigma}_t\boldsymbol{\epsilon}\mid \mathbf{x}_t=\mathbf{z}\right].
\end{equation}
A neural network $\mathbf{v}_\theta(\mathbf{x}_t,t)$ is trained with the standard velocity-matching objective
\begin{equation}
\mathcal{L}_{\mathrm{vel}}(\theta)=
\mathbb{E}_{\mathbf{x}_0,\boldsymbol{\epsilon},t}
\left[
\left\|\mathbf{v}_\theta(\mathbf{x}_t,t)-(\dot{\alpha}_t\mathbf{x}_0+\dot{\sigma}_t\boldsymbol{\epsilon})\right\|^2
\right].
\end{equation}

At inference time, samples are generated by numerically integrating \Cref{ode-eq} backward from $\mathbf{x}_T\sim\mathcal N(\mathbf{0},\mathbf{I})$ to $t=0$.

In practice, ODE solvers (e.g., Euler or diffusion ODE solvers such as DPM-Solver \cite{lu2022dpm}) evaluate the learned field $\mathbf{v}_\theta(\mathbf{x}_t,t)$ repeatedly across timesteps, and these network evaluations dominate inference cost. Hence, the number of function evaluations (NFEs) largely determines latency. Our goal is to reduce this cost by reusing denoiser outputs when a local change/error criterion indicates the update direction has changed only minimally.

\section{Sensitivity-Aware Caching}
In diffusion generative models, inference proceeds via an iterative denoising process over many timesteps that progressively removes noise to reconstruct the target sample. This procedure is expensive: each step is a full forward pass of a large network $f_\theta(\mathbf{x}_t, t, c)$, often repeated hundreds of times per sample. Caching seeks to avoid redundant computation by identifying steps where the network’s prediction changes only marginally and reusing a previously computed denoiser output. The central question is when reuse is safe. We seek a decision rule that, given the current inputs (latent $\mathbf{x}_t$, timestep $t$, and condition $c$), detects steps with negligible output change so the network evaluation can be safely skipped.

\subsection{Model Sensitivity}
Network sensitivity \cite{rifai2011contractive} quantifies how much a model’s output changes in response to small perturbations in its input. Formally, given a network $f_\theta(\mathbf{w})$, the local sensitivity is expressed through the Jacobian norm:
\begin{equation}
S(\mathbf{w}) = \left\| \frac{\partial f_\theta(\mathbf{w})}{\partial \mathbf{w}} \right\|,
\end{equation}
which measures how perturbations propagate through the network. Network sensitivity has been used to improve stability in deep networks \cite{glorot2010understanding,goodfellow2014explaining} and to analyze the robustness and smoothness of learned mappings \cite{rifai2011contractive}.

Intuitively, sensitivity captures how ``stiff'' or ``smooth'' the network function is around a given input.
A small Jacobian norm indicates a locally flat region where the network output varies little under small perturbations, 
whereas a large norm reveals a highly responsive or nonlinear regime. Thus, network sensitivity provides a principled basis for caching, as it identifies regions where the network output is locally less responsive to perturbations.

\begin{figure*}[t]
    \centering
    \begin{subfigure}[b]{0.6\textwidth}
        \centering
        \includegraphics[width=\textwidth]{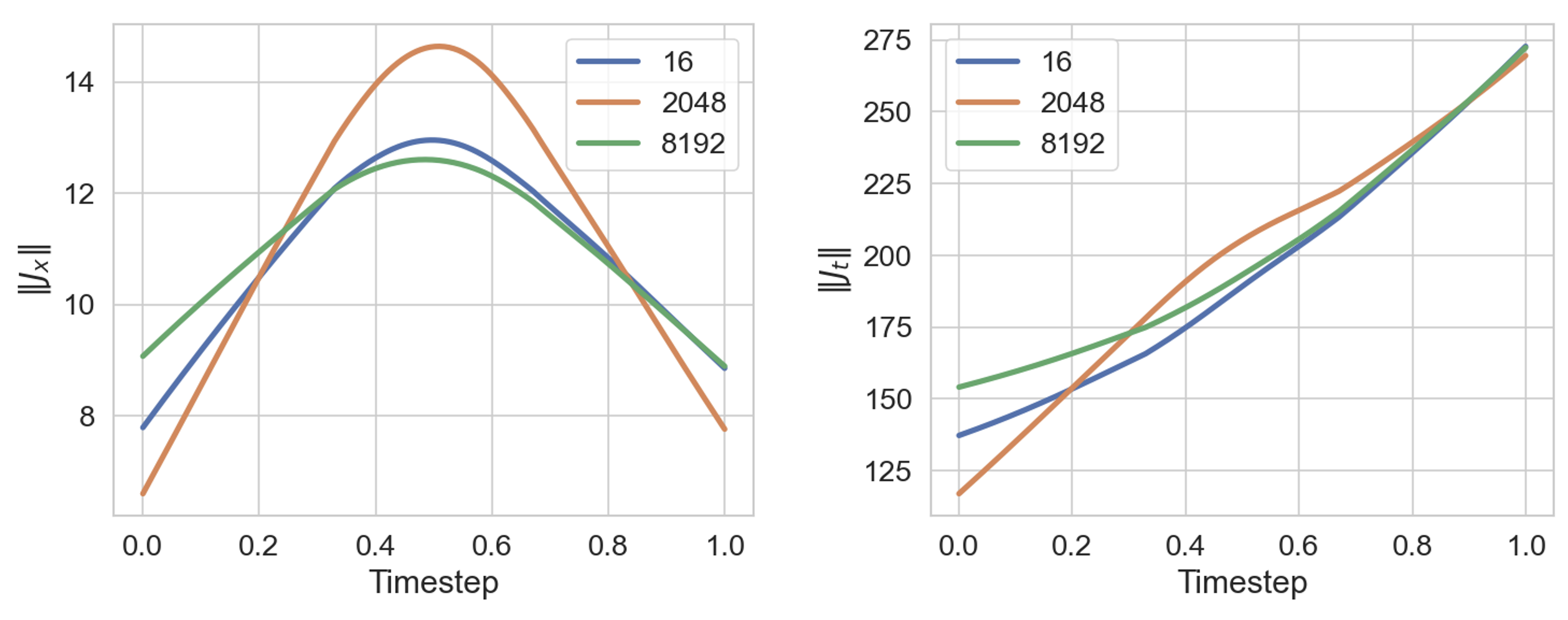}
        \caption{\textbf{Network Sensitivity Analysis.} (Left) Norm of the Jacobian w.r.t. the noisy latent. (Right) Norm of the Jacobian w.r.t. the timestep.}
        \label{fig:jacob-sitsubfig1}
    \end{subfigure}
    \hfill
    \begin{subfigure}[b]{0.36\textwidth}
        \centering
        \includegraphics[width=0.8\textwidth]{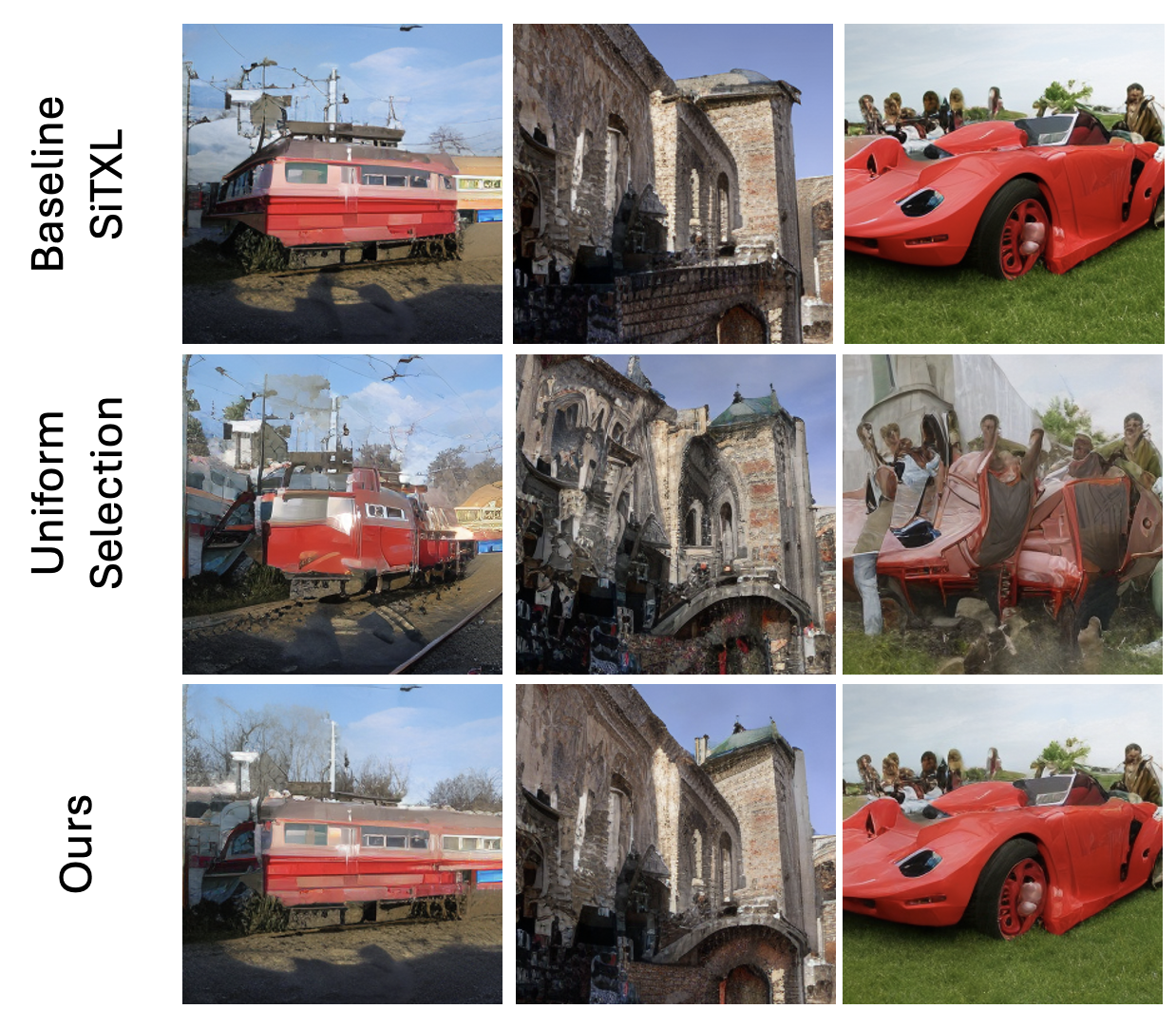}
         \caption{\textbf{Comparison of 25-step sampling} between Sensitivity-guided selection vs. Uniform selection.}
        \label{fig:jacob-sitsubfig2}
    \end{subfigure}
    \caption{\textbf{Sensitivity analysis of SiT-XL/2.} (a) We analyze the network's output sensitivity by computing the norm of the Jacobian with respect to the noisy latent (Left) and the timestep (Right). We observe that both inputs are significant for estimating changes in the network output. Furthermore, we find that this norm can be accurately approximated with a small number of samples; our comparison shows that 16 samples provide an estimate comparable to that from 2048 or 8192 samples, indicating that large batch sizes are not required for this estimation. (b) Leveraging this sensitivity score, we select an optimized subset of 25 denoising steps from a 250-step SDE sampler, compared against a baseline of uniform step selection. The sensitivity-guided method strategically skips steps where the network output exhibits low sensitivity (i.e., is not changing much), allowing for effective caching without harming output quality. The visual results demonstrate that samples generated with our method suffer minimal degradation, whereas the uniform selection baseline results in significant visual degradation.}
    \label{jacob-sit}
\end{figure*}

\subsection{Observation}
To probe when cached reuse is safe, we analyze the \emph{input sensitivity} of the SiT-XL/2 checkpoint \cite{ma2024sit} trained on ImageNet \(256\times256\) \cite{deng2009imagenet}.
For each sample and timestep \(t\), we compute the Jacobian and partial derivative of the denoiser output with respect to the noisy latent and the timestep, respectively:
\begin{equation}
J_x \;=\; \frac{\partial f_\theta(\mathbf{x}_t, t, c)}{\partial \mathbf{x}_t},
\qquad
J_t \;=\; \frac{\partial f_\theta(\mathbf{x}_t, t, c)}{\partial t},
\end{equation}
and record their norms \(\|J_x\|\) and \(\|J_t\|\).
We then aggregate these norms across validation samples and visualize their evolution as a function of \(t\) (see \Cref{jacob-sit}).

Two empirical observations emerge:
\begin{enumerate}
    \item \textbf{Non-negligible timestep sensitivity.}
    \(\|J_t\|\) attains consistently large values over wide ranges of \(t\), indicating that the network’s prediction is timestep-sensitive even when the latent change \(\|\Delta \mathbf{x}_t\|\) is small.
    Consequently, skipping across large \(\Delta t\) can incur noticeable error; caching purely on latent changes risks artifacts.
    \item \textbf{Both latent and timestep matter.}
    \(\|J_x\|\) is also substantial and varies with \(t\), implying that output deviation depends jointly on the latent drift \(\|\Delta \mathbf{x}_t\|\) and the timestep gap \(|\Delta t|\).
    Effective caching criteria therefore must account for both terms rather than relying on a single proxy.
\end{enumerate}

These findings motivate a sensitivity-based caching algorithm that explicitly combines latent and timestep changes, aligning cache decisions with the model’s local stability properties.

\subsection{Adaptive Sensitivity-Aware Caching}
The key challenge is to define a principled region of input change within which the denoiser’s output can be safely reused.

We quantify local output variation via \emph{input sensitivity}. A first-order expansion between consecutive steps gives:
\begin{equation}
f_\theta(\mathbf{x}_{t+\Delta t},\, t+\Delta t,\, c) - f_\theta(\mathbf{x}_t,\, t,\, c)
\;\approx\; J_x\,\Delta \mathbf{x}_t \;+\; J_t\,\Delta t,
\end{equation}
where $\Delta \mathbf{x}_t = \mathbf{x}_{t+\Delta t} - \mathbf{x}_t$. Taking norms yields the bound:

\begin{multline}
\big\| f_\theta(\mathbf{x}_{t+\Delta t}, t+\Delta t, c) - f_\theta(\mathbf{x}_t, t, c) \big\| \\
\le \|J_x\|\,\|\Delta \mathbf{x}_t\| + \|J_t\|\,|\Delta t|
+ \mathcal{O}\!\left(\|\Delta \mathbf{x}_t\|^2 + |\Delta t|^2\right).
\label{eq:sensitivity-bound}
\end{multline}

Thus, the Jacobian norms act as local Lipschitz constants governing the model responsiveness to latent and timestep perturbations.

We define the \emph{sensitivity score}
\begin{equation}
S_t \;=\; \|J_x\| \, \|\Delta \mathbf{x}_t\| \;+\; \|J_t\| \, |\Delta t|,
\label{eq:sensitivity-score}
\end{equation}
and adopt the following cache rule:
\begin{equation}
\text{Cache at step } t \;\;\Longleftrightarrow\;\; S_t \le \varepsilon,
\end{equation}
where $\varepsilon>0$ controls the accuracy–speed trade-off. When this criterion is met, the predicted change in output between $t$ and $t{+}\Delta t$ is below tolerance, so we reuse the cached $f_\theta(\mathbf{x}_t, t, c)$ instead of evaluating the network. \Cref{alg:sensitivity_caching} summarizes our proposed caching method.

\begin{algorithm}[t]
\caption{Sensitivity-Aware Caching}
\label{alg:sensitivity_caching}
\begin{algorithmic}[1]
\Require Denoiser $f_\theta$; tolerance $\varepsilon$; max cache length $n$; timesteps $\{t_k\}_{k=0}^{K}$; sampler; sensitivity cache $\mathcal{C}$
\State \textbf{Input:} $(\mathbf{x}_K,t_K,c)$
\State $y_K \gets f_\theta(\mathbf{x}_K,t_K,c)$
\State $(\mathbf{x}^r,t^r,y^r) \gets (\mathbf{x}_K,t_K,y_K)$
\State $(\alpha_x,\alpha_t) \gets \textsc{LookupSensitivity}(\mathcal{C}, t^r)$
\State $\mathbf{d} \gets \mathbf{0}$; $\tau \gets 0$; $m \gets 0$
\For{$k=K$ down to $1$}
    \State Obtain $(\mathbf{x}_{k-1}, t_{k-1})$ and $(\Delta \mathbf{x}_{k-1},\Delta t_{k-1})$ from sampler
    \State $\mathbf{d} \gets \mathbf{d} + \Delta \mathbf{x}_{k-1}$; $\tau \gets \tau + \Delta t_{k-1}$; $m \gets m+1$
    \State $S \gets \alpha_x \|\mathbf{d}\| + \alpha_t |\tau|$
    \If{$S \le \varepsilon$ \textbf{and} $m < n$}
        \State $y_{k-1} \gets y^r$ \Comment{cache hit}
    \Else
        \State $y_{k-1} \gets f_\theta(\mathbf{x}_{k-1}, t_{k-1}, c)$
        \State $(\mathbf{x}^r,t^r,y^r) \gets (\mathbf{x}_{k-1}, t_{k-1}, y_{k-1})$
        \State $(\alpha_x,\alpha_t) \gets \textsc{LookupSensitivity}(\mathcal{C}, t^r)$
        \State $\mathbf{d} \gets \mathbf{0}$; $\tau \gets 0$; $m \gets 0$
    \EndIf
\EndFor
\State \Return $\{y_k\}_{k=0}^{K}$
\end{algorithmic}
\end{algorithm}

\subsection{Relation to Prior Caching Methods}
\label{sec:relation-prior}

Our sensitivity view clarifies why prior heuristic policies can work and when they fail.

\paragraph{TeaCache.}
TeaCache~\citep{liu2025teacache} constructs step-skipping rules by modeling output residuals with differences in the time embedding. In our notation, this signal predominantly tracks changes along the timestep dimension, i.e., it approximates the $\|J_t\|\,|\Delta t|$ term in \Cref{eq:sensitivity-score}.
When the latent drift is small, focusing on the time embedding difference is reasonable. However, if the sampler induces a non-negligible change in the noisy latent (large $\|\Delta \mathbf{x}_t\|$), TeaCache underestimates the output change because it does not explicitly weight the $\|J_x\|\,\|\Delta \mathbf{x}_t\|$ contribution. This explains the artifacts observed when skipping across steps where the latent moves notably.

\paragraph{MagCache.}
MagCache~\citep{ma2025magcache} triggers skips based on the magnitude ratio of successive residual outputs. In our framework, this mainly reflects the $\|J_x\|\,\|\Delta \mathbf{x}_t\|$ component: small residual magnitudes typically indicate a locally gentle response to latent perturbations. The limitation is complementary to TeaCache: MagCache does not explicitly account for the timestep term $\|J_t\|\,|\Delta t|$, so it can be overconfident when the schedule takes larger $\Delta t$ steps or in regions where the denoiser is highly $t$-sensitive. 

Additionally, the first-order change between consecutive steps depends on $\Delta \mathbf{x}_t$ and $\Delta t$, not on $\Delta c$ (which is zero). This aligns with empirical observations reported by MagCache that caching quality is mostly independent of the prompt content, once $c$ is held fixed.

\subsection{Practical Implementation}
Since computing exact sensitivities is expensive, we approximate them using directional finite-difference (secant) estimates. More specifically, keeping $t$ fixed, the sensitivity with respect to $\mathbf{x}_t$ is estimated by:
\begin{equation}
\label{eq:Sx}
\|J_x\|
\;\approx\;
\frac{\bigl\| f_\theta(\mathbf{x}_t+\Delta \mathbf{x},\, t, c) - f_\theta(\mathbf{x}_t,\, t, c) \bigr\|_2}
     {\bigl\| \Delta \mathbf{x} \bigr\|_2},
\end{equation}
where $\Delta \mathbf{x}$ is a small perturbation in the solver step direction. Keeping $\mathbf{x}_t$ fixed, the sensitivity with respect to time is:
\begin{equation}
\label{eq:St}
\|J_t\|
\;\approx\;
\frac{\bigl\| f_\theta(\mathbf{x}_t,\, t + \Delta t, c) - f_\theta(\mathbf{x}_t,\, t, c) \bigr\|_2}
     {|\Delta t|}.
\end{equation}

These sensitivity values are computed once per model on a small calibration set and cached for use during inference. In our experiments, we use only 8 videos with varied motion dynamics and scene content for estimating sensitivity.

As the first-order estimation is only locally accurate, we introduce a hyperparameter $n$ that limits the maximum number of consecutive caching steps. 
After $n$ reuses, the cache is refreshed to prevent drift as the trajectory evolves. This parameter balances a trade-off between speed and accuracy: smaller $n$ yields conservative but stable caching, whereas larger $n$ provides higher speedups at the cost of reduced precision when the first-order approximation becomes inaccurate.

\section{Experiment}
\label{sec:experiment}

\begin{table*}[t]
\centering
\caption{We conduct a quantitative evaluation of inference efficiency and visual quality in video generation models. Efficiency was measured by the NFE and the cache ratio, while visual quality was assessed using LPIPS, PSNR, and SSIM. Our results show that with the same amount of compute as previous methods, SenCache achieves superior visual quality with improved LPIPS, PSNR, and SSIM scores.}
\label{tab:main}
\setlength{\tabcolsep}{6pt}
\renewcommand{\arraystretch}{1.12}
\begin{tabular}{lcccccc}
\toprule
\multirow{2}{*}{\textbf{Method}} &
\multicolumn{2}{c}{\textbf{Efficiency}} &
\multicolumn{4}{c}{\textbf{Visual Quality}} \\
\cmidrule(lr){2-3} \cmidrule(lr){4-7}
& \textbf{NFE $\downarrow$} & \textbf{Cache Ratio (\%) $\uparrow$} & \textbf{LPIPS $\downarrow$} & \textbf{PSNR $\uparrow$} & \textbf{SSIM $\uparrow$} \\
\midrule
\multicolumn{7}{l}{\textbf{Wan 2.1}~\cite{wan2025open} (\(T{=}50\)) \quad (81 frames, \(832\times480\))} \\
\midrule
TeaCache-slow & 33 & 34$\%$& 0.0384 & 30.6097 & 0.9372 \\
MagCache-slow & 25 & 50$\%$& 0.0390 & 30.7226 & 0.9387 \\
SenCache-slow  & 25 & 50$\%$& 0.0390 & 30.7233 & 0.9387 \\
\hdashline
TeaCache-fast & 25 & 50$\%$ &  0.0966 & 25.0661 & 0.8697 \\
MagCache-fast & \textbf{21} & \textbf{58$\%$} &  0.0603 & 28.3684 & 0.9143 \\
SenCache-fast & \textbf{21} & \textbf{58$\%$} &   \textbf{0.0540} & \textbf{29.1400} & \textbf{0.9219} \\
\midrule
\multicolumn{7}{l}{\textbf{CogVideoX}~\cite{yang2025cogvideox} (\(T{=}50\)) \quad (49 frames, \(720\times480\))} \\
\midrule
TeaCache            & \textbf{22} & \textbf{56$\%$}&  0.5855 & 14.0194 & 0.5702 \\
MagCache            & 23 & 54$\%$  &   0.1952 & 21.8502 & 0.7332 \\
SenCache           & \textbf{22} & \textbf{56$\%$}& \textbf{0.1901}  & \textbf{22.09} & \textbf{0.7786} \\
\midrule
\multicolumn{7}{l}{\textbf{LTX-Video}~\cite{HaCohen2024LTXVideo} (\(T{=}50\)) \quad (161 frames, \(768\times512\))} \\
\midrule
TeaCache            & 32 & 36$\%$&   0.2763 & 20.2019 & 0.7270 \\
MagCache            & 28 & 44$\%$& 0.1795 & 23.3655 & 0.8224 \\
SenCache            & \textbf{27} & \textbf{46$\%$}& \textbf{0.1625} & \textbf{23.6660} & \textbf{0.8293} \\
\bottomrule
\end{tabular}
\end{table*}

\begin{table*}[h]
\centering
\caption{\textbf{Ablation study on $n$, the maximum number of consecutive cache steps.} Performed on the Wan 2.1 model~\cite{wan2025open} with $\varepsilon = 0.05$. Increasing $n$ improves efficiency (lowers NFE) up to $n=4$, where NFE saturates. Further increasing $n$ provides no efficiency benefit and degrades visual quality, as the underlying finite-difference approximation becomes less accurate.}
\label{tab:ablation-n}
\begin{tabular}{lccccccc}
\hline
 & \textbf{$n = 1$} & \textbf{$n = 2$}  & \textbf{$n = 3$} & \textbf{$n = 4$} & \textbf{$n = 5$} & \textbf{$n = 6$} & \textbf{$n = 7$} \\
\hline
\textbf{NFE $\downarrow$ }& 32 & 27 & 25 & 23 & 23 & 23 & 23\\
\textbf{LPIPS $\downarrow$} & 0.0223 &0.0351 & 0.0454 &   0.0558 & 0.0636 & 0.0708 & 0.0760 \\
\textbf{PSNR $\uparrow$} & 32.8154 & 30.2874 &28.9923  &  28.0890 &27.4195 & 26.9355 & 26.5327 \\
\textbf{SSIM $\uparrow$} & 0.9583 &0.9419 &0.9301 & 0.9195 &0.9120& 0.9044 & 0.8991\\
\hline
\end{tabular}
\end{table*}

\begin{table}[h]
\centering
\caption{Ablation study on the error tolerance $\varepsilon$. Performed on Wan 2.1~\cite{wan2025open} with $n=3$. Results reveal a clear accuracy-efficiency trade-off.}
\label{tab:ablation-e}
\resizebox{\columnwidth}{!}{%
\begin{tabular}{lccccc}
\hline 
& \textbf{$\varepsilon = 0.04$}  & \textbf{$\varepsilon = 0.06$} & \textbf{$\varepsilon = 0.07$} & \textbf{$\varepsilon = 0.1$} & \textbf{$\varepsilon = 0.13$}  \\
\hline
\textbf{NFE $\downarrow$} & 25 & 23 & 22 & 22 & 21  \\
\textbf{LPIPS $\downarrow$} & 0.0455 & 0.0472 & 0.0485 & 0.0490 & 0.0513  \\
\textbf{PSNR $\uparrow$} & 29.0054 & 28.9292 & 28.9218 & 28.8320 & 28.7213  \\
\textbf{SSIM $\uparrow$} & 0.9301 & 0.9287 & 0.9277 & 0.9268 & 0.9244 \\
\hline
\end{tabular}%
}
\end{table}

\begin{figure}[t]
    \centering

    \includegraphics[width=\linewidth]{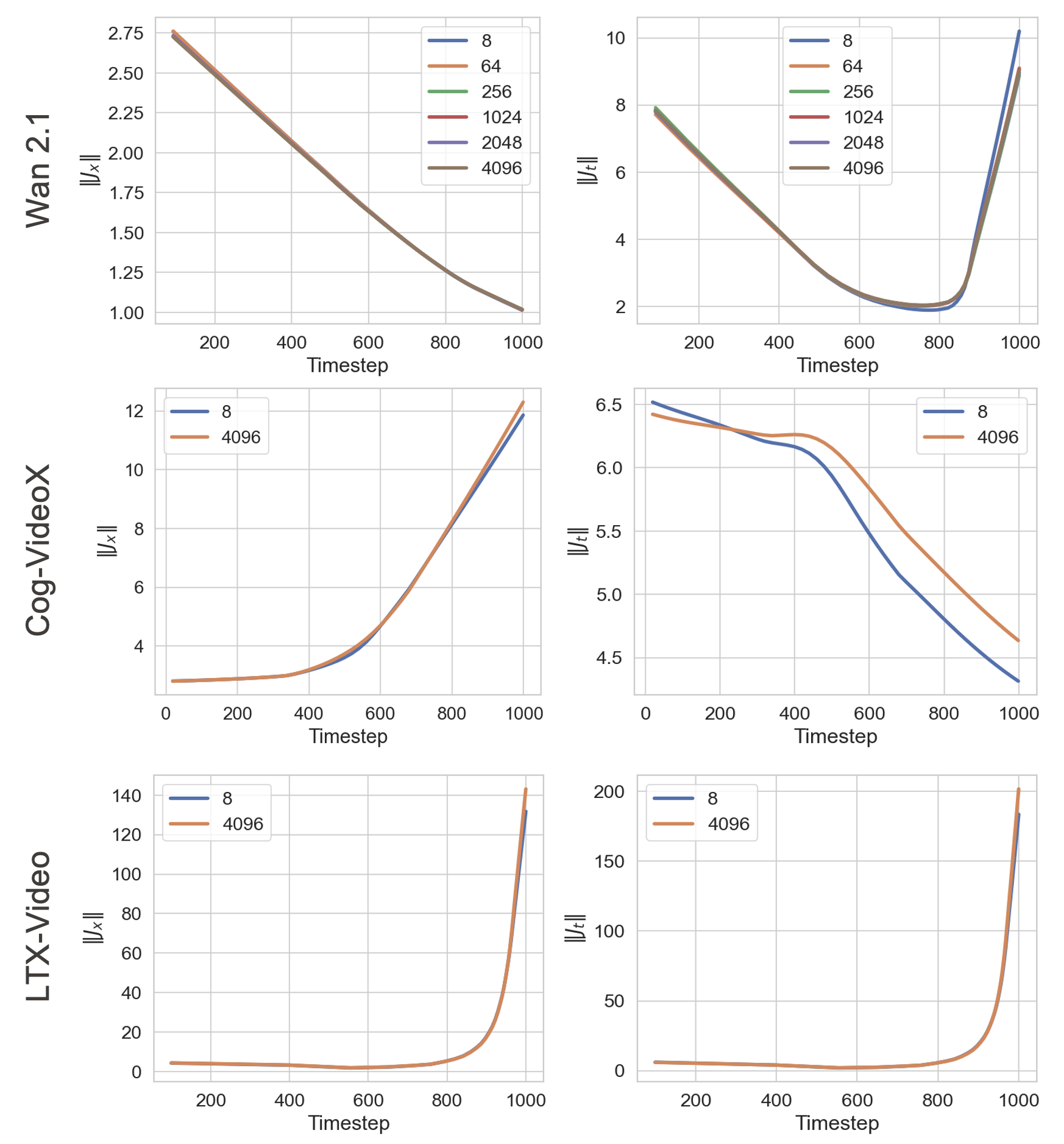}
    
    \caption{\textbf{Effect of calibration set size on sensitivity estimation.} We compare sensitivity profiles estimated from 8 videos versus 4096 videos and find that 8 diverse videos already yield a close match, indicating that large calibration sets are not required.}
    \label{fig:sensitivity-all}
\end{figure}

\paragraph{Settings.}
To demonstrate the effectiveness of our approach, we conduct quantitative evaluations on three state-of-the-art video diffusion models: Wan 2.1~\citep{wan2025open}, CogVideoX~\citep{yang2025cogvideox}, and LTX-Video~\citep{HaCohen2024LTXVideo}. We compare our method with TeaCache~\citep{liu2025teacache} and MagCache~\citep{ma2025magcache}. Following MagCache~\citep{ma2025magcache}, we report LPIPS~\citep{zhang2018lpips}, SSIM~\citep{wang2004ssim}, and PSNR as metrics for visual quality. We also report NFE (number of function evaluations) and Cache Ratio (percentage of denoising steps retrieved from cache) to assess computational efficiency.

\paragraph{Implementation details.}
We calculate the finite Jacobian norm estimate using 8 videos from the MixKit dataset~\citep{lin2024open}. We evaluate all methods on the full prompt set of VBench~\citep{huang2024vbench}. For the ablation study, we use 70 prompts randomly selected from T2V-CompBench (10 per category) to generate videos~\citep{sun2025t2vcompbench}. We use the hyperparameters from the official implementations of prior work~\citep{liu2025teacache,ma2025magcache}. For our method, we set $n=2$ for the slow version and $n=3$ for the fast version. As shown by previous work, the first $20\%$ denoising steps are critical to the overall generation process~\citep{ma2025magcache}; thus, we set a strict threshold of $1\%$ error ($\varepsilon = 0.01$) for these early steps. For the rest of the experiments, we set $\varepsilon = 0.1$ for Wan slow and $0.2$ for Wan fast, $0.6$ for CogVideoX, and $0.5$ for LTX.
\subsection{Main Results}
Our quantitative results are summarized in \Cref{tab:main}. We begin with Wan~2.1~\cite{wan2025open}. In the \emph{slow} (conservative) regime, all three methods achieve comparable visual quality, but TeaCache~\citep{liu2025teacache} yields lower speedups (higher NFE) than both MagCache~\citep{ma2025magcache} and SenCache. This suggests that when reuse is conservative, different reuse criteria often select similar ``safe'' regions of the denoising trajectory, and the remaining differences mainly affect the achieved compute reduction. In the \emph{fast} (aggressive) regime, the gap between methods becomes clearer: SenCache consistently attains better visual quality than MagCache at comparable compute (matched NFE), indicating that sensitivity-based decisions better identify timesteps where reuse incurs minimal degradation.

For CogVideoX~\citep{yang2025cogvideox} and LTX-Video~\citep{HaCohen2024LTXVideo}, matching the low NFE achieved by prior caching methods requires using larger tolerance values (e.g., $\varepsilon=0.5$ and $0.6$), corresponding to more permissive reuse. Under these aggressive settings, all methods exhibit a clearer quality drop, reflected by degraded LPIPS/PSNR/SSIM compared to Wan~2.1. Empirically, this indicates that CogVideoX and LTX-Video are less tolerant to approximation in the denoising updates, whereas Wan~2.1 appears to admit more reuse while preserving fidelity. Nevertheless, across all models and compute regimes, SenCache consistently achieves equal or better visual quality than prior caching baselines at similar (and often lower) NFE.

\subsection{Ablation Studies}
\paragraph{Ablation on $n$.}
We first ablate the cache lifetime parameter $n$, which limits the maximum number of consecutive cache reuses before a refresh. We run this study on Wan~2.1, fixing $\varepsilon=0.05$ and varying $n$. Results are summarized in \Cref{tab:ablation-n}. As expected, increasing $n$ reduces the number of function evaluations (NFE) by permitting longer reuse chains. Interestingly, the NFE improvement saturates beyond $n=4$: while the achieved NFE remains nearly unchanged, visual quality degrades. This suggests that overly long consecutive reuse chains can be harmful, as the first-order approximation becomes less accurate as the trajectory drifts away from the reference point.

\paragraph{Ablation on $\varepsilon$.}
We next ablate the tolerance $\varepsilon$ on Wan~2.1, fixing $n=3$ and varying $\varepsilon$ (see \Cref{tab:ablation-e}). We observe a clear accuracy--efficiency trade-off. Increasing $\varepsilon$ from $0.04$ to $0.13$ decreases NFE from $25$ to $21$, indicating that a more permissive threshold enables more aggressive caching. This reduction in compute comes with a gradual loss in fidelity: LPIPS increases from $0.0455$ to $0.0513$, PSNR drops from $29.01\,\mathrm{dB}$ to $28.72\,\mathrm{dB}$, and SSIM declines from $0.930$ to $0.924$. Over the tested range, the trends are approximately linear, supporting the interpretation of $\varepsilon$ as a tolerance that directly controls the reuse rate and thus the speed--quality trade-off. Notably, $\varepsilon\in[0.06,0.07]$ captures most of the NFE savings ($25\!\rightarrow\!22$--$23$) while incurring only minor quality degradation.

\paragraph{Ablation on calibration set size.}
Finally, we study how many videos are needed to obtain stable sensitivity estimates. We compute the sensitivity profiles using calibration sets ranging from 8 videos to 4096 videos and compare the resulting profiles (see \Cref{fig:sensitivity-all}). We find that using as few as 8 diverse videos yields sensitivity estimates that closely match those obtained with much larger calibration sets, suggesting that the sensitivity statistics are stable and that large calibration batches are not necessary in practice.
\section{Discussions and Future Work}
Currently, our implementation relies on a first-order sensitivity surrogate; an interesting direction would be to find efficient yet richer (higher-order or learned) estimators that could reduce error in nonlinear regimes. Additionally, as the sensitivity threshold $\varepsilon$ maps directly to an error budget at each denoising step, dynamically scheduling $\varepsilon$ across timesteps could further accelerate inference while maintaining generation quality: different steps contribute unequally to final fidelity, so allowing larger error at less critical stages may be acceptable. In this paper, we used a fixed threshold; designing schedules and characterizing effective patterns is left for future work. Finally, although we validated sensitivity-aware caching on video diffusion models, the core principle is not limited to the visual domain. Extending this approach to other modalities—such as text, audio, or multimodal diffusion systems—represents an exciting avenue for future research.

\section{Conclusion}
\label{sec:conclusion}
In this work, we introduced a principled framework to accelerate diffusion model inference by leveraging the local smoothness of the denoising network. 
By quantifying model sensitivity with respect to both the noisy latent and the timestep, we developed a principled criterion for deciding when cached outputs can be safely reused. 
Our analysis revealed that both latent and temporal sensitivities play critical roles in determining the validity of cache reuse, motivating a combined metric that adapts to the network’s local behavior. 
We further proposed an efficient finite-difference approximation to estimate these sensitivities in practice, requiring only a small calibration set and a single precomputation per model. 
Experiments on video diffusion models demonstrated that this strategy significantly reduces inference cost while maintaining generation quality. We hope this sensitivity-based perspective can serve as a foundation for future adaptive acceleration methods across diffusion architectures and modalities.

\section*{Acknowledgments}
This work was supported as part of the Swiss AI Initiative
by a grant from the Swiss National Supercomputing Centre
(CSCS) under project ID a144 on Alps. It has also been supported by Swiss National Science Foundation (SNSF) under Grant No. 10003100.

{
    \small
    \bibliographystyle{ieeenat_fullname}
    \bibliography{main}
}

\begin{figure}[h]
    \centering
    \includegraphics[width=\linewidth]{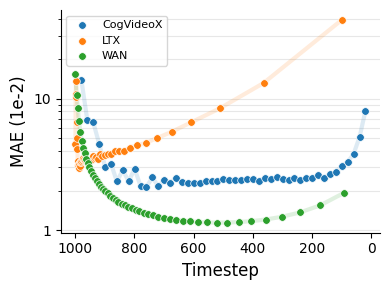}
    \caption{MAE between the denoiser outputs at two consecutive timesteps.}
    \vspace{-0.5cm}
    \label{fig:diff}
\end{figure}

\clearpage
\setcounter{page}{1}
\maketitlesupplementary
\paragraph{Insight on CogVid and LTX high $\varepsilon$.} We design the following diagnostic. For each of the three models, we generate 100 videos and compute the mean absolute error (MAE) between the denoiser outputs at two consecutive timesteps, i.e.,
$\|f(x_{t_k},t_k)-f(x_{t_{k-1}},t_{k-1})\|_1$.
Smaller values indicate that reusing a cached output across nearby steps would introduce less error. The averages over 100 videos are reported in \Cref{fig:diff}. We observe that in the mid-range timesteps (approximately 800--200), where caching is most frequently applied, this consecutive-step MAE is consistently higher for CogVideoX and LTX-Video than for Wan. This suggests these models exhibit larger per-step variation (higher effective sensitivity), so achieving the same NFE reduction requires a larger caching tolerance $\varepsilon$, which inherently permits more approximation error and can lead to the observed quality drop.

\paragraph{SenCache vs Global Timestep Optimization Methods.}
Local sensitivity in SenCache is a proxy for the marginal cost of skipping one more step. Global schedule methods can be seen as doing the same thing but with planning: they allocate the error budget across timesteps to avoid cases where many “small” local skips add up. In this view, SenCache is like using a fixed per-step budget (via $\varepsilon$), while global optimization is the more general version that chooses how that budget should vary over time. An interesting future direction is to combine the two: a global scheduler could provide dynamic $\varepsilon(t)$ values that SenCache uses for local decisions.

\paragraph{Additional Efficiency Metrics.} On a GH200 GPU for Wan 2.1, our method reduces end-to-end wall-clock latency from $182.3$\,s (vanilla) to $107.3$\,s ($41.1\%$ speedup), compared to MagCache at $110.6$\,s ($39.3\%$ speedup); both reduce total compute from $8{,}244{,}043.09$ to $3{,}482{,}412.58$ GFLOPs ($57.8\%$ fewer).

\paragraph{Cross-Model Sensitivity Patterns.}
We provide a visualization of the estimated network sensitivity in \Cref{fig:sensitivity-all}. Across all models, we first observe that variations in both the time step $t$ and the noisy sample must be taken into account for effective caching, as the networks are sensitive to both. Second, a small batch of $8$ diverse samples is already sufficient to obtain reliable sensitivity estimates; large batches are not necessary. Third, the sensitivity patterns differ markedly between models. For Wan 2.1  and LTX, at large timesteps, the model is highly sensitive to variations in $t$. However, this is not the case for CogVideoX. Moreover, while CogVideoX and LTX exhibit low sensitivity to input variations at small timesteps, wan 2.1 shows the opposite behavior and is highly sensitive in this regime. Finally, for LTX in particular, we observe that it is highly sensitive to both variation in $t$ and the noisy latent at large timesteps, but is less sensitive at smaller time steps.

\end{document}